\documentclass{article}
\usepackage{arxiv}

\usepackage{times}
\usepackage{soul}
\usepackage{url}
\usepackage[hidelinks]{hyperref}
\usepackage[utf8]{inputenc}
\usepackage[small]{caption}
\usepackage{graphicx}
\usepackage{amsmath}
\usepackage{amsthm}
\usepackage{booktabs}
\usepackage{float}
\usepackage{tcolorbox}
\usepackage[ruled,vlined]{algorithm2e}
\usepackage{researchpack}
\usepackage{wrapfig}
\usepackage{natbib}
\usepackage[capitalize,nameinlink]{cleveref} % noabbrev
\usepackage{enumitem}
\usepackage{wrapfig}
\usepackage{xspace}
\usepackage{multirow}
\usepackage{color,soul}
\usepackage{subcaption}
\usepackage{graphicx}
\usepackage{caption}
\usepackage[dvipsnames]{xcolor}

\usepackage{tikz}
\usetikzlibrary{shapes,arrows}

\hypersetup{
    colorlinks=true,
    linkcolor=blue,
    citecolor=blue,
    urlcolor=blue
}

\usepackage{pifont}

\definecolor{plotgreen}{HTML}{00DC00}

\usepackage{booktabs}
\newcommand{\tradeverse}{\textsc{TradeVerse} }

\title{TradeVerse: A Longitudinal Benchmark of Political Negotiation in International Trade}

\date{} 					% Or removing it

\author{Debodeep Banerjee\\
	DI, University of Pisa\\
        DISI, University of Trento\\
	\And
	  Amitangshu Dasgupta  \\
	   amitangshudasgupta@gmail.com  \\
}

\renewcommand{\shorttitle}{\tradeverse}

\begin{document}

\maketitle

\begin{abstract}
LLMs are increasingly being applied to tasks involving institutional and political texts, but existing benchmarks evaluate them on isolated documents or single tasks. In realpolitik, negotiations are longitudinal data, where participating parties can align or argue over multiple iterations and each turn is an outcome of the previous turns, hence, understanding one turn requires tracking everything before it. We introduce \tradeverse, a benchmark built from the World Trade Organisation (WTO) specific trade concerns, where member states challenge one another and exchange arguments over multiple rounds, sometimes for years. We, in \tradeverse, reconstruct  minutes of $1170$ meetings, spanning across $5$ groups and $89$ product groups and define three tasks: first, the system has to analyze the longitudinal meeting records and predict the harmonized system codes (HS chapters) of the products under discussion in the particular meeting, second, we examine whether the system, upon analyzing the anonymized content of the meeting, can guess the name of the responding country and third, we ask the system to play the role of the responding country and provide the statement for the very last round. All labels are recovered directly from the proceedings, requiring no manual annotation. Our experiments highlight the challenges these tasks pose for current LLMs. To the best of our knowledge, \tradeverse is the first benchmark to investigate potential of LLMs in understanding longitudinal political trade negotiations. 
\end{abstract}


\section{Introduction}
% \begin{itemize}
%     \item negotiation benchmarks
%     \item political benchmark
%     \item none of them evaluate if an LLM can successfully follow a negotiation
    
% \end{itemize}

% \DB{Write your comment}
Large language models (LLMs) are increasingly being used in legal, political and institutional decision making tasks \citep{bignotti2024legal, nangia2026unsc, liang2026benchmarking, nguyen2025llms}. Existing benchmarks for such high-stake domains predominantly evaluate LLMs on static decision-making or document understanding tasks, whereas many real-world institutional decisions emerge through negotiations that evolve over multiple interactions. Such scenarios are complex and dynamic as the conflicting parties may change their stances over time, or may even involve other parties in order to support them. Therefore, such dialogues may not be always bidirectional and the LLM, while performing any task based on such data, must take into account the arguments of all participating parties. More importantly, unlike traditional reasoning benchmarks where the input is a static document, the notion of longitudinality arises through the successive interaction of multiple parties centering around a single agenda. In this regard, successful reasoning requires a language model to interpret the inherent dynamics of conversation put forward by the participating parties. 

\begin{figure*}[!t]
    \centering
    \includegraphics[width=1.0\linewidth]{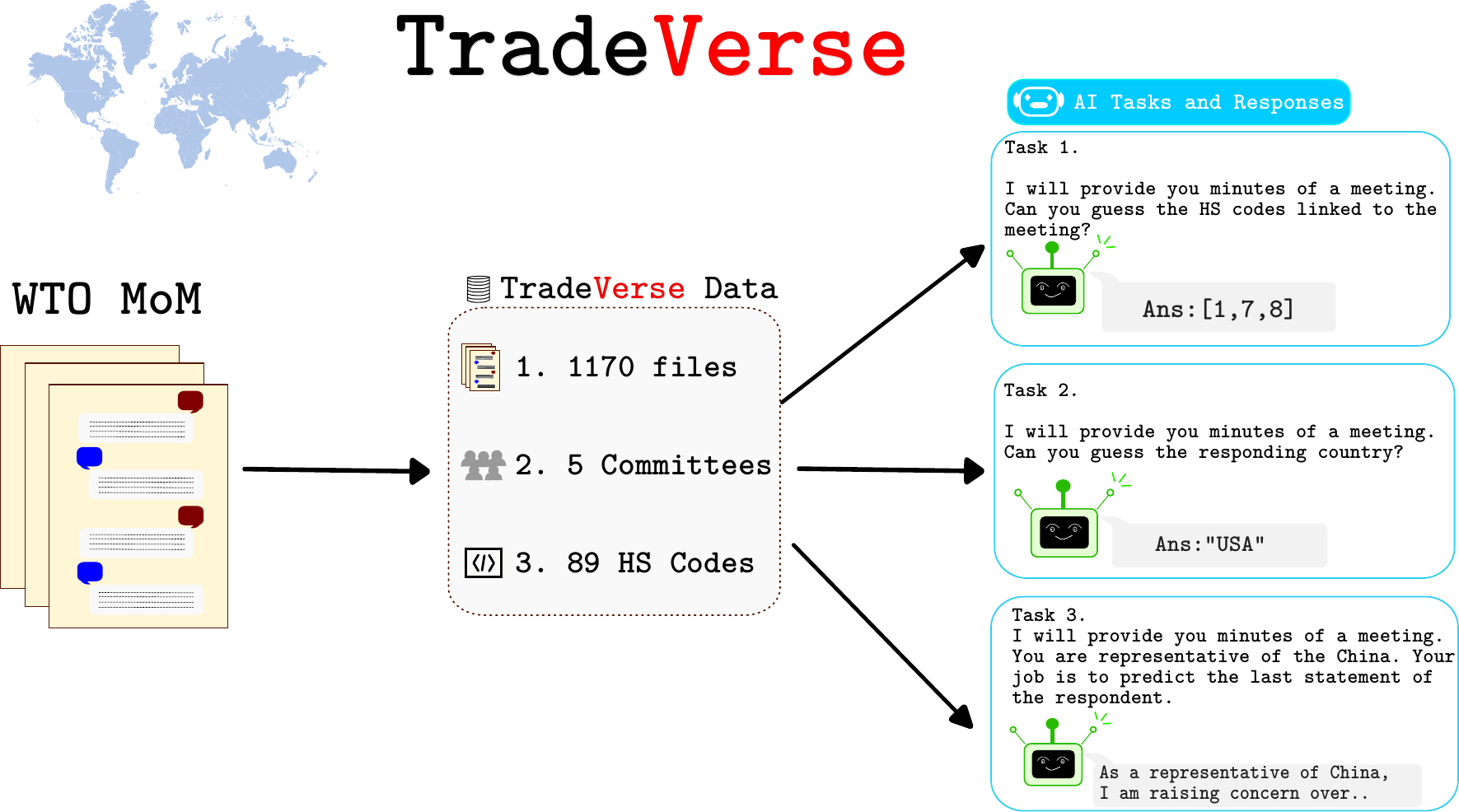}
    \caption{Overview of the \tradeverse benchmark pipeline. These streams feed into three core evaluation tasks: predicting product categories (Task 1), identifying responding members under anonymization (Task 2), and generating the concluding statement of the respondent (Task 3).}
    \label{fig:overview}
\end{figure*}

International trade dialogues offer a challenging setting for such politically motivated longitudinal texts. In academic literature, the World Trade Organisation (WTO) is largely viewed as a novel solution to settle trade disputes \citep{lang2009hidden}.  The minutes of the meetings (MoM) released on the trade concerns \footnote{\url{https://tradeconcerns.wto.org/en}} documented in the WTO database is a perfect example of such negotiations. In such scenarios, two or more countries engage in dialogues based on certain goods or products. The arguments are inherently based on the country's vested interests, making the content rich in political contexts. Such trade disputes are seldom singleton meetings as the concerned state(s) may raise same complaint several times, seek support from other countries. The responding country, similarly engages in the dialogue to put forward its arguments -- stretching the meetings to several rounds, sometimes spanning years. Therefore, in such cases, the LLMs should take into account the country in action, its supporting members along with the main content of the arguments.

% **Unlike parliamentary debates or legislative proceedings, WTO trade concerns are structured institutional negotiations with clearly defined participant roles (raisers, supporters, and respondents), making them particularly well suited for evaluating reasoning over evolving multi-party interactions.**

%AD{** ?}
In this paper, we propose \tradeverse, a longitudinal political benchmark based on the WTO trade-concern MoMs. \tradeverse reconstructs $1170$ MoMs, spanning across $5$ WTO groups and $89$ product codes (HS chapters). The structure of these negotiations offers three complementary reasoning problems. First, prediction of the HS chapters linked to a particular meeting, second, prediction of the \textit{responding} country, and third, generating the last statement for the responding country. Note that, all the three tasks mentioned above test the LLM's ability to understand the chronologies of the political dialogues. Furthermore, none of the tasks requires human annotations as the ground truths can be readily extracted from the MoMs. An abridged version of an actual dialogue between China and T\"urkiye is provided in \Cref{fig:example}. The full conversation is provided in the appendix. 

Through these tasks, \tradeverse evaluates whether current LLMs can move beyond surface-level document understanding and reason over the temporal, institutional, and strategic structure of international negotiations. Our experiments show that models identify responding countries with high accuracy, but this accuracy is unevenly distributed: all six models we evaluate are substantially more accurate for Western members than for the rest of the membership. Models further over-predict product categories, achieving high recall at low precision, and generate final statements that are fluent and diplomatically appropriate yet share little specific content with the interventions actually made. More broadly \tradeverse provides a benchmark for studying LLMs reasoning under evolving institutional processes, geopolitical biases and strategic language generation in authentic international discussion. All three aforementioned tasks are crucial to analyze a language model's real potential in the field of realpolitik.   

To summarise, our contributions are: 
\begin{itemize}
    \item We introduce \tradeverse{}, a longitudinal benchmark of authentic
    multi-party WTO trade negotiations spanning five groups and more than
    three decades of institutional proceedings.

    \item We reconstruct $1{,}170$ trade concerns from $6{,}933$ meeting records
    and represent each concern as a structured sequence of interventions by
    raiser, supporter, and respondent members.

    \item We define three annotation-free tasks---HS chapter prediction,
    respondent identification, and final-statement generation with ground
    truths recovered directly from official WTO proceedings.

    \item We evaluate six contemporary LLMs and identify systematic limitations,
    including over-prediction of product categories, limited specificity in
    generated diplomatic responses, and a respondent-identification disparity
    favouring Western members that persists under full anonymization.

    \item We release the benchmark dataset, extraction pipeline, and evaluation
    code to support further research on longitudinal, institutional, and
    geopolitical reasoning.
\end{itemize}

\begin{figure}[!t]
\centering
\begin{tcolorbox}[colback=blue!3, colframe=blue!40!black, boxrule=0.5pt,
                  arc=2pt, left=5pt, right=5pt, top=4pt, bottom=4pt,
                  fonttitle=\bfseries, fontupper=\normalsize,
                  title={T\"urkiye --- Additional tariffs on electric vehicles (Market Access Committee, ID~100)}]

\textbf{Meeting 1 (Apr 2023)}\\
\textit{China (raiser):} ``\ldots T\"urkiye sharply increased import tariffs on China-made electric vehicles \ldots the bound rate \ldots is 20\%. The import tariff \ldots has reached 50\% \ldots''\\[3pt]
\textit{T\"urkiye (respondent):} ``We would like to thank China for its interest in this issue, which is currently being discussed \ldots in respective Capitals.''

\medskip
\textbf{Meeting 2 (Oct 2023)}\\
\textit{China (raiser):} ``China regrets to raise this issue again. \ldots a presidential decree imposing a 40\% additional tariff only on \ldots electric vehicles originating from China.''\\[3pt]
\textit{T\"urkiye (respondent):} ``The electric vehicles sector, as an infant industry, has a strategic importance for T\"urkiye \ldots T\"urkiye, within its bound rates, increased its MFN tariff rate \ldots''

\medskip
\textbf{Meeting 3 (Mar 2024)}\\
\textit{China (raiser):} ``\ldots China is highly concerned about this measure. \ldots China urges T\"urkiye to correct its wrongdoing \ldots immediately.''\\[3pt]
\textit{T\"urkiye (respondent):} \colorbox{yellow!30}{\strut\textbf{[Task 3 target --- withheld]}}

\end{tcolorbox}
\caption{A representative \tradeverse{} concern across three meetings (abridged). China raises the same concern
repeatedly while T\"urkiye's position develops from acknowledgement to a
substantive infant-industry defence. The product (Task~1) and responding member
(Task~2) are inferred from the exchange; the respondent's final statement
(Task~3) is generated from the preceding history.}
\label{fig:example}
\end{figure}

\section{Related works}
\paragraph{Reasoning benchmarks for large language models.}

%AD{Further, Nangia, Gokrani, and Lazzaroni (2026) proposed UNSC-Bench, where LLMs are prompted to assume the roles of specific countries and predict the voting decisions of the five permanent members of the UN Security Council on proposed resolutions.}
The rapid progress of LLMs has motivated the development of various benchmarks with increasing difficulties. While benchmarks such as MMLU-Pro \citep{wang2024mmlu}, Big-bench \cite{ghazal2013bigbench}, MMLU-global \citep{singh2025global} examine LLMs across diverse range of domains, reasoning problems, languages, and cultural settings, complementary benchmarks evaluate LLMs' capabilities of reason under long contexts. \citet{bai2023longbench} introduced Longbench, a bilingual long-context multi-task benchmark. \citet{kuratov2024babilong} introduced Babilong, another long context benchmark. The benchmark includes 20 reasoning tasks where the facts are scattered across the long context. However, while such benchmarks caused great advancements in evaluating the potentials of LLMs, such benchmarks fall short in testing the LLMs abilities in understanding political nuances, in particular, evaluating LLM's potential in understanding longitudinal political texts, where discourse of meetings on the same issue can evolve over multiple iteration.

\paragraph{Political benchmarks.}
Recent work has begun to evaluate LLMs in authentic political and institutional settings. \citet{liang2026benchmarking} introduced UNBench, an unified corpus based on four tasks, spanning from drafting to statement generation. Further \citet{nangia2026unsc} proposed UNSC-Bench, where the LLM is prompted to play the role of certain countries to predict the resolutions taken by the UN permanent members. A benchmark on the European Parliament (EU) database was proposed by \citet{zhang2025policon}. Beyond benchmarks, LLMs bias in tasks related to political science was studied by \citet{choi2026eastern}. However, none of these benchmarks essentially addresses what \tradeverse is advocating, these benchmarks does not follow whether the model is required to follow the course of action over different meetings. Although \citet{liang2026benchmarking} introduces a statement generation task, it is worth noting that at the UN, a statement is made after the votes. 

\paragraph{LLMs in trade.}
Besides general purpose and political benchmarks, substantial research has taken place in analysing LLMs' roles in trade. Nevertheless, such applications are primarily restricted in evaluating LLM's role in financial discourse. \citet{qian2026agents} introduced Agent Market Arena (AMA), a lifelong real-life benchmark. Besides benchmarks, LLMs potential has been investigated under diverse ranges of financial task \citep{chen2025stockbench,xiao2024tradingagents, lopez2025can}. However, these models typically address tasks related to financial, falling short in evaluating quality of a real-world trade negotiation. \citet{mahajan2025llms}, however, acknowledge this gap and propose TradeGov, a question-answer based dataset related to 5k trade laws involving 138 countries.  The dataset is created with ChatGPT \citep{achiam2023gpt}. Although TradeGov is the closest related resource, it focuses exclusively on legal questions. To the best of our knowledge, ours is the first dataset explicitly curated to address this issue systematically.
 
\paragraph{Our contribution.}
TradeVerse complements these research directions by introducing a benchmark built from the World Trade Organization (WTO) Trade Concerns Database, consisting of 1170 longitudinal negotiations spanning five WTO groups and 89 product groups. Unlike existing benchmarks, \tradeverse focuses on how well the language model can follow the evolution of a conversation among countries. Together, these tasks require models to integrate long-context understanding, institutional reasoning, structured prediction, and natural language generation within authentic multilateral trade negotiations.

\section{\protect\tradeverse}
\label{sec:tradeverse}
The \tradeverse benchmark is constructed from the WTO Trade Concerns Database. The Trade Concerns Database is a chronological collection of dialogues that take place among WTO member states regarding specific trade concerns. An illustrative description is provided in \Cref{fig:overview}.

The database is organized into five groups. Within each committee, multiple meetings are documented. Each meeting consists of one or more \emph{raiser states} and one or more \emph{respondent states}. In addition,
the raiser state may receive support from one or more \emph{supporter states}. However, the presence of supporter states is not mandatory for every meeting.

We define the components of our dataset as follows.

Let $\mathcal{D}=\{d_1,d_2,\ldots,d_n\}$ denote the set of all meeting minutes, $\mathcal{C}=\{c_1,\ldots,c_m\}$ be the set of WTO member states, and $\mathcal{H}=\{h_1,\ldots,h_k\}$ be the set of Harmonized System (HS) codes.

For each meeting minute $d_i,\qquad i\in\{1,\ldots,n\}$, there exist $R_i$ negotiation rounds, where $R_i \ge 2$. That is, we only consider trade concerns that have been discussed over at least two rounds of meetings. We denote the rounds of concern $d_i$ as $r_{i,1}, r_{i,2}, \ldots, r_{i,R_i}$, where each round $r_{i,j}$ comprises the statements made by the raiser, supporter, and respondent states at the $j$-th meeting of the concern. 

Based on this dataset, we define the following benchmark tasks.

\paragraph{Task 1: HS Code Prediction}
The harmonized system codes (HS chapters) is conventionally used to classify import and export goods \footnote{\url{https://www.trade.gov/harmonized-system-hs-codes}}. Using HS codes, various goods are numerically classified in a hierarchical order. There are 97 principal chapters for the codes and each chapter is subdivided in multiple branches. However, in this research, we only ask the models to predict the HS chapters. Given the full negotiation history of a trade concern, the objective is to predict the set of Harmonized System (HS) codes associated with it. Formally,
\[
\hat{h}_i
=
\mathcal{L}\!\left(
r_{i,1}, r_{i,2}, \ldots, r_{i,R_i}
\right),
\qquad \hat{h}_i \subseteq \mathcal{H},
\]
where $\hat{h}_i$ denotes the predicted set of HS chapters and $\mathcal{L(\cdot}$ is the LLM in use. As a concern may involve multiple products, this is a multi-label prediction task.

\paragraph{Task 2: Respondent Prediction}
Given the negotiation history with all country identities masked, the objective is to identify the responding state. We consider two settings of increasing difficulty. In the first, only the responding country is masked (ST-1); in the second, all country names are masked (ST-2). Let $\tilde{r}_{i,j}$ denote round $r_{i,j}$ under the
applicable masking. While under ST-1, only the respondent country along with its demonyms and national agencies, is replaced by a placeholder token, ST-2 incorporates placeholder tokens to every mention of a member state along with its demonyms and national agencies. The task is
\[
\hat{c}_i
=
\mathcal{L}\!\left(
\tilde{r}_{i,1}, \tilde{r}_{i,2}, \ldots, \tilde{r}_{i,R_i}
\right),
\qquad \hat{c}_i \in \mathcal{C},
\]
where $\hat{c}_i$ is the predicted respondent state. As the identity of every participant is hidden, the model must infer the respondent from the substance of the dispute rather than from explicit references.

\paragraph{Task 3: Statement Generation}
Given the negotiation history preceding the final round, together with the raiser and supporter statements of the final round, the objective is to generate the respondent's concluding statement. Let $r_{i,R_i}^{\setminus \text{resp}}$ denote the final round with the respondent's statement removed, and let $s_i$ denote that withheld statement. The task is defined as, 
$
\hat{s}_i
=
\mathcal{L}\!\left(
r_{i,1}, \ldots, r_{i,R_i-1},\; r_{i,R_i}^{\setminus \text{resp}}
\right),
$
where $\hat{s}_i$ is the generated statement, evaluated against the reference $s_i$.

% Write me!
% \section{Dataset details}
\paragraph{Data collection.}
Data have been collected from the WTO trade concern database, a public record of concerns raised by member states against other member states. Each concern is deliberated within one of five groups and may be discussed among multiple iterations, sometimes even for years. To this end, we noticed that only meetings belonging to 2 of these 5 groups contain HS chapters.

We collect the data in two phases. Once the raw data was collected, we preprocessed the data for the three aforementioned tasks.

First, upon collection of the metadata, we obtain the structured metadata consisting of the meeting IDs, committee, member countries with different roles, number of rounds, etc. To maintain the longitudinal notion of the dataset, we discarded all meetings that have less than two meetings.

For each of the remaining meetings, we first downloaded the documents of the minutes. The documents are Word documents, and the country names are written in different colours. We parsed these documents to organise countries and their roles and corresponding statements.

In our final version of the dataset, we focus on the countries, their roles, statements, and the numbers of rounds used in a particular meeting.

\begin{table}[t]
\centering
\caption{\tradeverse{} corpus statistics.}
\label{tab:dataset_stats}
\begin{tabular}{lr}
\toprule
\textbf{Statistic} & \textbf{Value} \\
\midrule
Trade concerns              & 1{,}170 \\
Meeting records             & 6{,}933 \\
Statements                  & 26{,}219 \\
groups                  & 5 \\
Distinct responding members & 68 \\
HS chapters covered         & 89 \\
Meetings per concern (mean) & 5.9 \\
Meetings per concern (max)  & 52 \\
Year range                  & 1995--2026 \\
\bottomrule
\end{tabular}
\end{table}

\paragraph{Statistics and insights.}

\begin{figure}[htbp]
    \centering
    \includegraphics[width=0.9\linewidth]{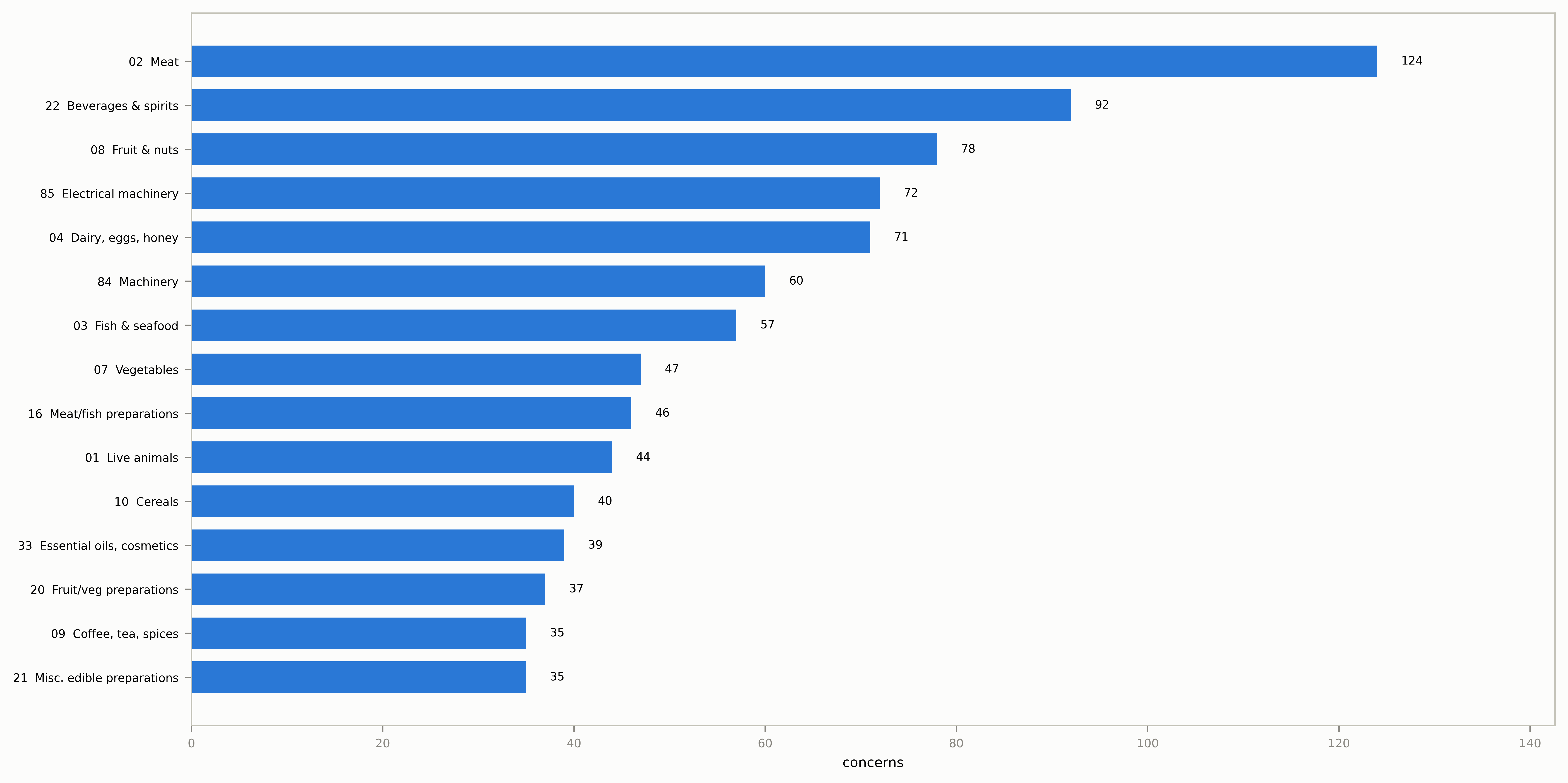}
    \caption{Frequency of top-15 HS chapters as per their frequency in the corpus.}
    \label{fig:hscodes}
\end{figure}

\begin{table*}[t]
\centering
\caption{Performance of different LLMs on HS code prediction.
Higher values indicate better performance.}
\label{tab:hs_code_results}
\begin{tabular}{lcccccc}
\toprule
\multirow{2}{*}{\textbf{Model}}
& \multicolumn{3}{c}{\textbf{Micro}}
& \multicolumn{3}{c}{\textbf{Macro}} \\
\cmidrule(lr){2-4}
\cmidrule(lr){5-7}
& \textbf{Precision}
& \textbf{Recall}
& \textbf{$F_1$}
& \textbf{Precision}
& \textbf{Recall}
& \textbf{$F_1$} \\
\midrule

Nemotron3-Ultra:550B
& 54.52 & 75.18 & 63.21
& 42.21 & 58.49 & 46.46 \\

Llama-3.3:70B-Instruct
& 37.76 & 63.50 & 47.36
& 13.17 & 23.84 & 15.68 \\

GLM-5.2
& 61.12 & 70.86 & \textbf{65.63}
& 46.98 & 51.91 & \textbf{46.55} \\

DeepSeek-V4-Pro
& 62.52 & 63.63 & 63.07
& 28.35 & 25.90 & 25.89 \\

GPT-OSS-120B
& 62.02 & 64.48 & 63.23
& 43.38 & 40.85 & 39.90 \\

Kimi-K2.7-Code
& 54.38 & 74.62 & 62.91
& 42.40 & 54.00 & 45.18 \\

\bottomrule
\end{tabular}
\end{table*}

\begin{figure}[!t]
    \centering
    \includegraphics[width=0.9\linewidth]{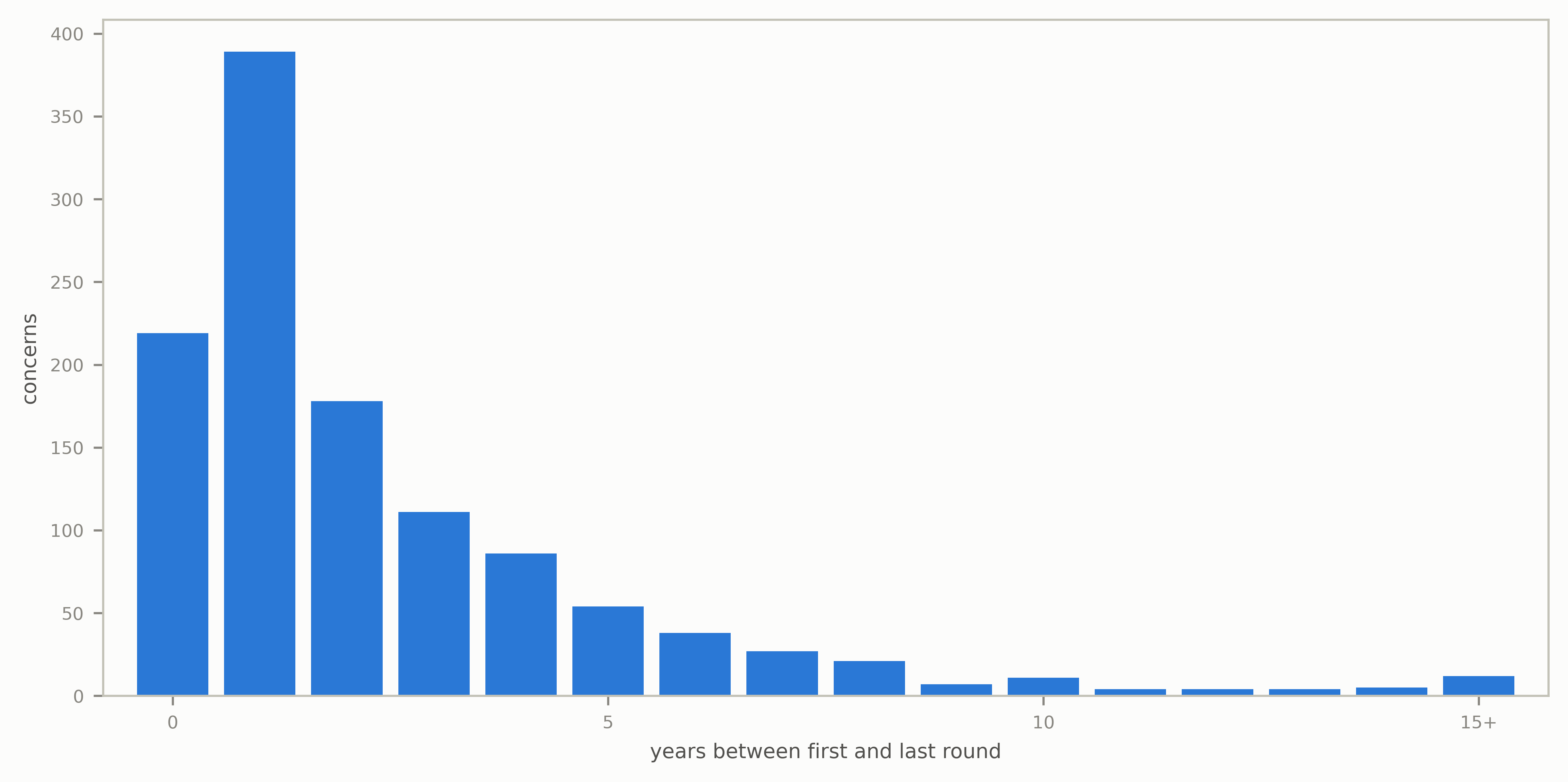}
    \caption{Distribution of concerns across total number of years it lasted for}
    \label{fig:years_disputes}
\end{figure}

% \AD{filtered lekha but ekhane mentioned nei je singleton round wala meetnings baad deowa hoyechhe, mention korle better, intro e mentioned achhe. reader k search/scroll backt korte hobe ki filter holo noeto.}
% \DB{Eta \tradeverse er surutei bola ache. $R_i\ge2$}
Our corpus contains a total of 1170 concerns, encompassing a total of 6933 meeting records. While the HS code system consists of 97 chapters globally, we observed that our dataset covers 89 chapters. The HS chapters are primarily dominated by agricultural foods and products, followed by electronic items (See \Cref{fig:hscodes}) .   
Individual concerns, typically ranging between $2-52$ meetings with a mean of 5.9. The concerns are distributed among 5 groups -- TBT (538), SPS (397), CTG (133), CMA (68), and IL (34). Summary statistics are reported in Table~\ref{tab:dataset_stats}. The corpus consists of a total of 68 parties as respondents with EU \footnote{As EU may appear along with other individual countries, we consider `EU' as a country in this research.} having the highest number of responses (224), followed by India (118). Our data spans across more than three decades (1995-2026) where several meetings can last for years. A distribution of years against number of concerns are presented in \Cref{fig:years_disputes} where the span between first and last round of a meeting  is reported against the total number of corresponding concerns. It is interesting to note that while the mean span of a concern is 2.5 years, few of the concerns last for more than 15 years with a maximum span of 20 years. 
\section{Experimental details}
% \begin{table*}[t]
% \centering
% \caption{Performance of different LLMs on HS code prediction.
% Higher values indicate better performance.}
% \label{tab:hs_code_results}
% \begin{tabular}{lcccccc}
% \toprule
% \multirow{2}{*}{\textbf{Model}}
% & \multicolumn{3}{c}{\textbf{Micro}}
% & \multicolumn{3}{c}{\textbf{Macro}} \\
% \cmidrule(lr){2-4}
% \cmidrule(lr){5-7}
% & \textbf{Precision}
% & \textbf{Recall}
% & \textbf{$F_1$}
% & \textbf{Precision}
% & \textbf{Recall}
% & \textbf{$F_1$} \\
% \midrule

% Nemotron3-Ultra:550B
% & 54.52 & 75.18 & 63.21
% & 42.21 & 58.49 & 46.46 \\

% Llama-3.3:70B-Instruct
% & 37.76 & 63.50 & 47.36
% & 13.17 & 23.84 & 15.68 \\

% GLM-5.2
% & 61.12 & 70.86 & \textbf{65.63}
% & 46.98 & 51.91 & \textbf{46.55} \\

% DeepSeek-V4-Pro
% & 62.52 & 63.63 & 63.07
% & 28.35 & 25.90 & 25.89 \\

% GPT-OSS-120B
% & 62.02 & 64.48 & 63.23
% & 43.38 & 40.85 & 39.90 \\

% Kimi-K2.7-Code
% & 54.38 & 74.62 & 62.91
% & 42.40 & 54.00 & 45.18 \\

% \bottomrule
% \end{tabular}
% \end{table*}
\subsection{Models}
We employ several SOTA LLMs including Nvidia Nemotron Ultra 550b \citep{blakeman2026nemotron}, DeepSeek-V4-Pro \citep{xu2026deepseek}, GLM-5.2 \citep{zeng2026glm}, Llama3.3:70b \citep{grattafiori2024llama}, GPT-OSS:120b \citep{agarwal2025gpt}, and Kimi-K2.7-Code \citep{team2025kimi} for the tasks mentioned earlier. While task 1 and task 2 are classification based, task 3 is related to natural language generation. We used Together API to run the models. For each of the task, the models are tested under a zero-shot setting at temperature 0.0. The prompts used for each task is provided in the appendix. 

\subsection{Metrics}
% \AD{Task ordering ta dekhish ekbar, 2nd task HS Code prediction lekha}
Each of the models are evaluated with a variety of metrics. The first task is about predicting the HS chapters and the model can produce more than one codes corresponding to the meeting -- hence, evaluating with \textit{precision}, \textit{recall} and $F1$ have been legitimate choices. However, for the first task, we evaluate the models' performance by measuring \textit{Accuracy}. As the third task is an NLG based task, we evaluate the models' performance both lexically and semantically. We used BLEU-4 \citep{papineni2002bleu} and ROUGE-L \citep{lin2004rouge} scores to check the lexical similarities and we checked the semantic similarities using bert score \citep{zhang2019bertscore}. We used A100 GPU and microsoft/deberta-large-mnli model \citep{he2021deberta} to calculate the bert score. 

\subsection{Evaluation}
\paragraph{Task~1.} Task~1 evaluates whether models can interpret a trade dialogue well enough to identify the products under discussion. Given a full discussion as input, the model predicts the set of HS chapters associated with it. Note that, Meetings related to only two groups, TBT and SPS have corresponding HS codes. This constraint led us to use 724 out of 1170 meetings to predict  the HS codes. Since a single concern may span several products, we impose no constraint on the number of codes generated. Table~\ref{tab:hs_code_results} reports the performance of six models. Across all models, recall substantially exceeds precision: models correctly recover the relevant chapters but over-generate, predicting additional chapters that are not associated with the concern. This suggests that models hedge under uncertainty, listing plausible neighbouring product categories rather than committing to a precise set.
\begin{table}[t]
\centering
\caption{Accuracy under models in Task 2 under two anonymization settings.}
\label{tab:masking_accuracy}
\scriptsize
\begin{tabular}{lcc}
\toprule
\textbf{Model} &
\textbf{ST-1 (\%)} &
\textbf{ST-2 (\%)} \\
\midrule
Nemotron3-Ultra:550B     & 90.68 & 90.60 \\
Llama-3.3:70B-Instruct    & 85.89 & 84.69 \\
GLM-5.2                         & \textbf{92.86} & \textbf{93.21} \\
DeepSeek-V4-Pro                 & 92.14 & 91.20 \\
GPT-OSS-120B                    & 82.63 & 80.92 \\
Kimi-K2.7-Code                  & 91.78 & 91.87 \\
\bottomrule
\end{tabular}
\end{table}

\begin{figure}[htbp]
    \centering
    \includegraphics[width=0.99\linewidth]{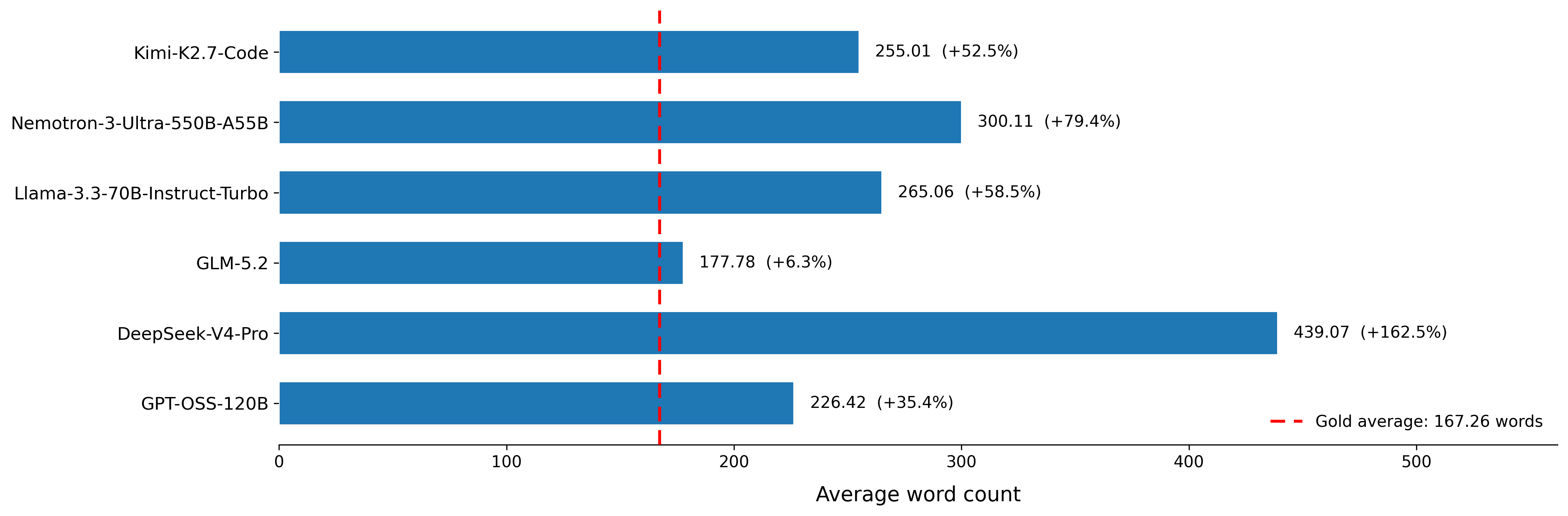}
    \caption{Comparison of average length of predictions against the average length of ground truth prediction. The \textcolor{red}{red dotted line} indicates the average word count of the ground truth responses.}
    \label{fig:avg_resp}
\end{figure}

\begin{table*}[!t]
\centering
\caption{Respondent identification accuracy for Western vs.\ non-Western members under both anonymization settings. Every difference is significant at $p<0.01$. The gap—Western minus non-Western accuracy—persists in both settings and, for several models, widens when all participant identities are masked.}
\label{tab:western_nonwestern}
% \scriptsize
\begin{tabular}{llccccc}
\toprule
\textbf{Setting} & \textbf{Model}
& \textbf{Western (\%)} & \textbf{Non-Western (\%)}
& \textbf{$\Delta$} & \textbf{$\chi^2$} & \textbf{$p$-value} \\
\midrule
\multirow{6}{*}{\rotatebox[origin=c]{90}{\textbf{ST-1}}}
& Nemotron3-Ultra:550B        & 97.74 & 87.05 & 10.69 & 34.28 & $0.00$ \\
& Llama-3.3:70B-Instruct  & 96.22 & 80.57 & 15.65 & 51.70 & $0.00$ \\
& GLM-5.2                       & 97.99 & 90.20 &  7.79 & 22.83 & $2\times10^{-6}$ \\
& DeepSeek-V4-Pro               & 97.99 & 89.12 &  8.87 & 27.31 & $0.00$ \\
& GPT-OSS-120B                  & 92.70 & 77.46 & 15.24 & 41.35 & $0.00$ \\
& Kimi-K2.7-Code                & 96.98 & 89.09 &  7.89 & 20.64 & $6\times10^{-6}$ \\
\midrule
\multirow{6}{*}{\rotatebox[origin=c]{90}{\textbf{ST-2}}}
& Nemotron3-Ultra:550B         & 98.24 & 86.66 & 11.58 & 40.02 & $0.00$ \\
& Llama-3.3:70B-Instruct  & 94.46 & 79.66 & 14.80 & 43.12 & $0.00$ \\
& GLM-5.2                       & 97.47 & 90.96 &  6.51 & 16.38 & $5\times10^{-5}$ \\
& DeepSeek-V4-Pro               & 96.98 & 88.21 &  8.77 & 24.09 & $1\times10^{-6}$ \\
& GPT-OSS-120B                  & 91.94 & 75.26 & 16.68 & 46.18 & $0.00$ \\
& Kimi-K2.7-Code                & 97.24 & 89.09 &  8.15 & 22.22 & $2\times10^{-6}$ \\
\bottomrule
\end{tabular}
\end{table*}

\paragraph{Task~2.} We investigate whether models can identify the responding country from an anonymized transcript. We consider two settings namely ST-1 and ST-2 as mentioned earlier. If models relied on the identities of the other participants, who is raising the concern, who is offering support to infer the respondent, removing all names should degrade performance substantially. Table~\ref{tab:masking_accuracy} shows this is not the case: accuracy is nearly identical across the two settings for every model, and in two cases (GLM-5.2, Kimi-K2.7) it is marginally higher when all names are hidden. This indicates that models do not rely on the surrounding participants' identities but instead infer the respondent from the substance of the dispute—the measures, products, and legal arguments at issue.
\begin{table}[htbp]
\centering
\caption{Performance of different LLMs on final-round response generation.
All results are reported as percentages. Higher values indicate better performance.}
\label{tab:generation_results}
\scriptsize
\begin{tabular}{lccccc}
\toprule
\textbf{Model}
& \textbf{BLEU-4}
& \textbf{ROUGE-L}
& \multicolumn{3}{c}{\textbf{BERTScore}} \\
\cmidrule(lr){4-6}
& &
& \textbf{Pr.}
& \textbf{Rc.}
& \textbf{$F_1$} \\
\midrule

Nemotron3-Ultra:550B
& 7.82
& 19.57
& 55.65
& 61.19
& 57.83 \\

Llama-3.3:70B-Instruct
& 6.86
& 18.09
& 53.85
& 60.29
& 56.67 \\

GLM-5.2
& \textbf{11.31}
& \textbf{23.00}
& 57.59
& 64.32
& \textbf{60.49} \\

DeepSeek-V4-Pro
& 8.91
& 21.05
& 58.64
& 61.39
& 59.71 \\

GPT-OSS-120B
& 2.91
& 14.31
& 48.19
& 60.00
& 53.23 \\

Kimi-K2.7-Code
& 7.49
& 18.82
& 56.04
& 60.57
& 57.81 \\

\bottomrule
\end{tabular}
\end{table}

\paragraph{Task~3} In this final task, we test how well the LLM can interpret the conversation by employing it to generate the statement of the respondent country. To this end, as mentioned earlier, the models get access to look at the statements provided by all the countries and is tasked with generating the statement of the respondent country for the very last round. We noticed that several documents, despite having more than 1 round, do not contain any statement from the respondent country or in case there is exactly 2 rounds, the response is missing in any of the rounds. As a result, we decided to remove such instances from our corpus and achieved 1101 out of 1170 meetings for task 3. As complete dependence on the lexical similarities is insufficient for such tasks, we also check the bert scores for semantic similarities. We report the model performances in \Cref{tab:generation_results}. It is clear that  GLM outperforms its competitors in both lexical and semantic similarities, showing the model's superiority in understanding and generating statements that are more coherent to the actual statements. Furthermore, in \Cref{fig:avg_resp} we notice that, while all the models overshoots average word count of the ground truth statements, GLM showcases the most conservative estimate with an average word count of 177.78 against the 167.26, the average word count of the ground truth. Note that, in this task, when the model can observe the statements generated in the previous rounds, it attains the opportunity to learn the notion of the conversation, in the sense the historical texts work as a training signal to the model -- eventually helping it playing its role as the respondent country. To this end, we also investigate whether the number of rounds have any significant impact on the models' performances. Therefore, the correlation between number of rounds for each document and its corresponding Bert-$F_1$ scores are calculated and reported in \Cref{tab:rounds_correlation}. The table suggests that all the $\rho$s except GPT-OSS are statistically significant at $1\%$ significance level and the rest is at $5\%$. The result confirms that the number of rounds can potentially impact the models performance. The spearman's $\rho$ reported in the table suggests that number of rounds has a positive correlation with the models' performances.

% \begin{figure}[htbp]
%     \centering
%     \includegraphics[width=0.65\linewidth]{AuthorKit27/Figures/average_response_word_counts.png}
%     \caption{Caption}
%     \label{fig:avg_resp}
% \end{figure}

\begin{table}[htbp]
\centering
\caption{Spearman correlation between the number of negotiation rounds and response quality (BERTScore-$F_1$).}
\label{tab:rounds_correlation}
\begin{tabular}{lcc}
\toprule
\textbf{Model} & \textbf{Spearman $\rho$} & \textbf{p-values}\\
\midrule
Nemotron3-Ultra:550B     & 0.210 & 0.00\\
Llama-3.3:70B-Instruct    & 0.205 & 0.00\\
GLM-5.2                         & \textbf{0.314} & 0.00\\
DeepSeek-V4-Pro                 & 0.304 & 0.00\\
GPT-OSS-120B                    & 0.070 & 0.02\\
Kimi-K2.7-Code                  & 0.207 & 0.00\\
\bottomrule
\end{tabular}
\end{table}

\section{Discussion}

Our central finding does not concern how the language models identify supporting countries. Rather, they identify respondent countries better. In order to investigate this, we separated the respondent countries into two cohorts—Western and non-Western blocs. Following \citet{pritchard2011comparing}, we classify Australia, Austria, Canada, Finland, France, Germany, Greece, Ireland, Italy, Japan, the Netherlands, New Zealand, Norway, Portugal, Spain, Sweden, Switzerland, the United Kingdom, and the United States as Western. We additionally classify the European Union as a Western supranational entity when it appears as a respondent in the WTO records. We present the outcome of this exercise in \Cref{tab:western_nonwestern}.

Notably, respondents representing the Western bloc are identified substantially more accurately across all six models evaluated in this research, with every difference significant at $p<0.01$, upon conducting a Chi-square test of independence. We tested whether models' correctness and the geopolitical region are independent. To this end, it is also evident that this particular trait is visible in both anonymization settings, highlighting the fact that the disparity is stable. In some cases, the gap ($\Delta$) is even larger when all the countries mentioned in the minutes are masked.

This indicates that the language models are focusing more on the nuances of the dialogue rather than looking for any hints from the country names. However, one tentative explanation for such results could be the models' exposure to varying amounts of training data based on the countries representing the Western bloc. In such a case, the model may be able to isolate the target country, backed by its internal knowledge.

Nonetheless, investigating the asymmetry in the models' performance is beyond the scope of this research, and we reserve it for future work.

\paragraph{Application to language modeling.}
\tradeverse offers a wide range of applications to the machine learning and language modelling community, stretching beyond what is presented in this research. Since there has been active research on understanding LLMs' capabilities in negotiation \citep{bianchi2024well,kwon2024llms,hua2024game}, TradeVerse appears to be a natural real-world application.

In this research, we have shown how a single LLM can be applied to solve a series of tasks. However, it would be interesting to investigate how multiple agents, each representing different countries and their respective roles, can communicate with each other.

\paragraph{Application to social scientists.}
% Beyond evaluating language models, \tradeverse{} may serve researchers in trade policy and the social sciences. The corpus is, to our knowledge, the first machine-readable reconstruction of WTO Specific Trade Concerns as structured multi-party dialogue, with speaker roles, products, and meeting histories recovered from the proceedings. Trade-policy scholars can use it to study patterns of regulatory contestation---which measures attract concerns, how coalitions form, and how disputes resolve over time---at a scale difficult to assemble by hand. For computational social science, it offers a corpus of real institutional negotiation for work on argumentation, coalition dynamics, and diplomatic language. Our own finding of a geopolitical disparity in model performance also bears on a methodological concern for these fields: as LLMs are increasingly used to analyze political text, systematic unevenness in how well they represent different actors is directly relevant to anyone applying them to under-studied regions or smaller states.

Beyond evaluating language models, \tradeverse may serve researchers in social sciences as well. To the best of our knowledge, the corpus is first reconstruction of WTO trade concern proceedings as a structured multi party dialogues, number of rounds, relevant product codes and respective domains. Such a corpus would help researchers to understand how disputes evolve over time, how coalition forms and patterns of participating countries. \tradeverse helps to conduct research in such directions, especially at a scale that is typically challenging to accumulate with human effort. To this end, the relevance of \tradeverse also applicable to computational social scientists as well. Researchers from this field can understand how institutional negotiations unfolds, or any particular pattern of diplomatic language. 

\section{Conclusion}
We introduced \tradeverse, a longitudinal benchmark constructed from the World Trade Organization's Specific Trade Concerns, in which member states contest one another's regulations across meetings held over months or years. From these proceedings we defined three tasks—HS chapter prediction, respondent identification, and final-statement generation, whose ground truth is recovered directly from the record, requiring no manual annotation. Evaluating six contemporary language models, we found that they identify responding members with high accuracy overall, but that this accuracy is systematically lower for members outside the Western group across every model tested, a gap that remains when all participant identities are masked and cannot be attributed to surface cues. \tradeverse{} offers a setting in which the institutional and geopolitical structure of real trade negotiation can be studied directly, and we release the dataset, extraction pipeline, and evaluation code to support further work.
\bibliography{reference}
\bibliographystyle{plainnat}

% \section{Appendix}
\appendix
\section{Appendix}
\subsection{Additional Data Analysis}
\label{app:analysis}
\begin{figure}[htbp]
    \centering
    \includegraphics[width=0.7\linewidth]{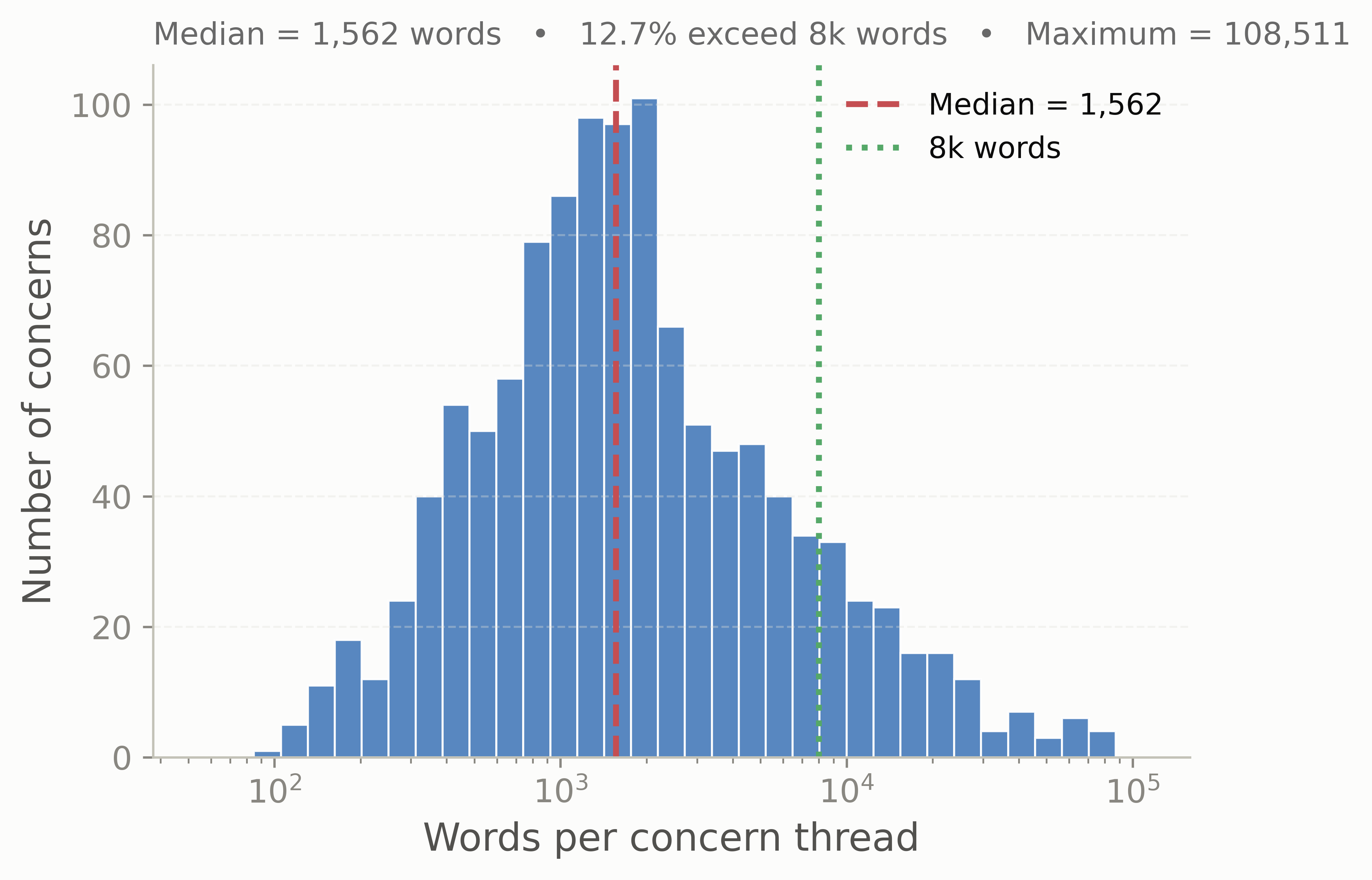}
    \caption{Mean word count vs number of concerns. Concern threads are heavy-tailed in length. }
    \label{fig:thread_words}
\end{figure}
\begin{figure*}[htbp]
    \centering
    \includegraphics[width=0.9\linewidth]{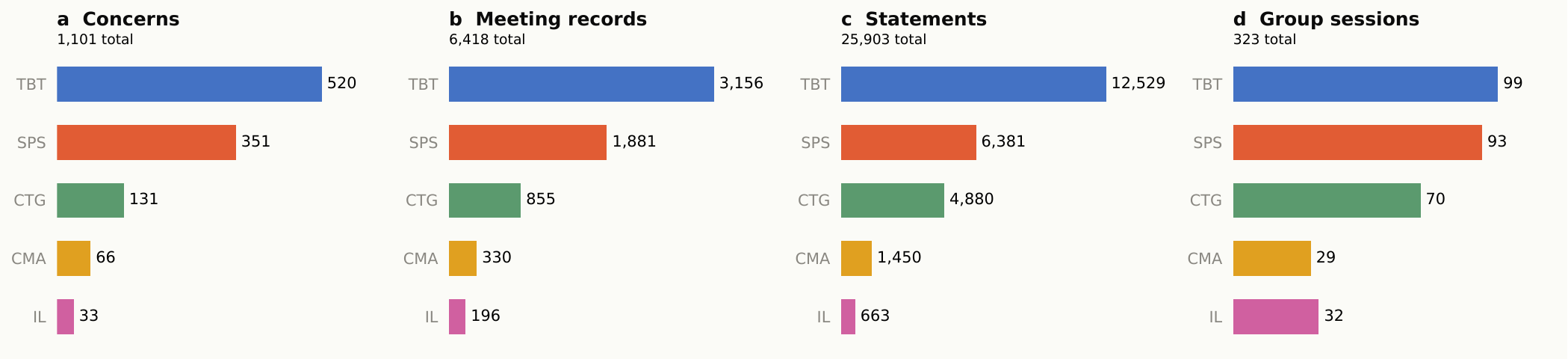}
    \caption{Distribution of the corpus across the five WTO groups. .}
    \label{fig:committees}
\end{figure*}

This section provides further descriptive analysis of the \tradeverse corpus. We first characterise the length of concern threads
(\Cref{fig:thread_words}), which motivates the long-context demands of the benchmark, and then examine how the corpus is distributed across the five WTO committees along several units of analysis (\Cref{fig:committees}). The x-axis is logarithmic and bins are log-spaced, so the visually near-symmetric shape is an approximately log-normal body with a long right tail: the median thread is 1,562 words (red dashed line), but the mean is over 4,500 and the longest thread runs to 108,511 words — a span of more than three orders of magnitude. The green dotted line marks 8,000 words; 12.7\% of threads exceed it, and the top percentile exceeds 50,000 words. Thread length is therefore a live constraint for the benchmark tasks: a model that must condition on a full concern history will fit most threads comfortably into a short context, but a non-trivial minority demand long-context handling or truncation, and those are disproportionately the multi-round, long-running concerns that the tasks are hardest on.
Each panel counts a different unit of analysis: (a) specific trade concerns, i.e. longitudinal threads; (b) meeting records, one per (concern, round) pair; (c) individual member statements; and (d) unique committee sessions, i.e. distinct sittings of a committee, at which many concerns are typically discussed. TBT, SPS, CTG, CMA, and IL, ordered by concern count and coloured consistently across all figures. Note that panels use independent horizontal scales, so bar lengths are comparable within a panel but not across panels. TBT and SPS together account for 79\% of concerns and 73\% of statements, but the corpus is far more evenly spread over sessions: Import Licensing contributes only 3\% of concerns (33) while accounting for 10\% of sessions (32), reflecting that its committee meets regularly but with a light concern docket per sitting. Counts are from the cleaned corpus (1,101 concerns, 6,418 meeting records, 25,903 statements, 323 sessions; 1995–2026)

% extra figures/tables
\subsection{Additional analysis on evaluation}
\begin{table*}[htbp]
\centering
\caption{Ablation on prior negotiation history for statement generation
(Task~3). We report the 95\% bootstrap confidence interval on the gain in
BERTScore-$F_1$ from providing the full history versus the final round alone,
the paired effect size (Cohen's $d$), and the Wilcoxon signed-rank $p$-value.
Prior history significantly improves generation for five of six models;
GPT-OSS-120B shows no measurable effect (CI includes zero).}
\label{tab:ablation}
\begin{tabular}{lcccc}
\toprule
\textbf{Model} & \textbf{$n$} & \textbf{95\% CI} & \textbf{Cohen's $d$} & \textbf{Wilcoxon $p$} \\
\midrule
GLM-5.2                  & 1101 & [0.041, 0.048]    & 0.75 & $<$0.001 \\
DeepSeek-V4-Pro          & 1100 & [0.028, 0.035]    & 0.61 & $<$0.001 \\
Llama-3.3-70B-Instruct   & 1100 & [0.016, 0.021]    & 0.44 & $<$0.001 \\
Nemotron-3-Ultra-550B    & 1099 & [0.020, 0.028]    & 0.35 & $<$0.001 \\
Kimi-K2.7-Code           & 1100 & [0.014, 0.020]    & 0.32 & $<$0.001 \\
\midrule
GPT-OSS-120B             & 1101 & [$-$0.000, 0.005] & 0.05 & 0.31 \\
\bottomrule
\end{tabular}
\end{table*}

\begin{table*}[!t]
\centering
\caption{Respondent identification accuracy for Western vs.\ non-Western members under both anonymization settings. The gap—Western minus non-Western accuracy—persists in both settings. Statistical significance is assessed using a chi-square test.}
\label{tab:western_nonwestern}
\begin{tabular}{llccccc}
\toprule
\textbf{Setting} & \textbf{Model}
& \textbf{Western (\%)} & \textbf{Non-Western (\%)}
& \textbf{$\Delta$} & \textbf{$\chi^2$} & \textbf{$p$-value} \\
\midrule

\multirow{6}{*}{\rotatebox[origin=c]{90}{\textbf{ST-1}}}
& Nemotron3-Ultra:550B      & 97.06 & 87.05 & 10.01 & 13.11 & $2.94\times10^{-4}$ \\
& Llama-3.3:70B-Instruct    & 94.12 & 80.57 & 13.55 & 17.19 & $3.4\times10^{-5}$ \\
& GLM-5.2                   & 97.06 & 90.20 &  6.86 & 7.52  & 0.0061 \\
& DeepSeek-V4-Pro           & 95.88 & 89.12 &  6.76 & 6.55  & 0.0105 \\
& GPT-OSS-120B              & 87.06 & 77.46 &  9.60 & 7.22  & 0.0072 \\
& Kimi-K2.7-Code            & 95.88 & 89.09 &  6.79 & 6.59  & 0.0103 \\

\midrule

\multirow{6}{*}{\rotatebox[origin=c]{90}{\textbf{ST-2}}}
& Nemotron3-Ultra:550B      & 96.47 & 86.66 &  9.81 & 12.17 & $4.86\times10^{-4}$ \\
& Llama-3.3:70B-Instruct    & 91.76 & 79.66 & 12.10 & 12.93 & $3.23\times10^{-4}$ \\
& GLM-5.2                   & 96.43 & 90.96 &  5.47 & 4.84  & 0.0278 \\
& DeepSeek-V4-Pro           & 93.53 & 88.21 &  5.32 & 3.55  & 0.0597 \\
& GPT-OSS-120B              & 84.71 & 75.26 &  9.45 & 6.49  & 0.0108 \\
& Kimi-K2.7-Code            & 95.29 & 89.09 &  6.20 & 5.39  & 0.0203 \\

\bottomrule
\end{tabular}
\end{table*}

\paragraph{Geopolitical disparity in respondent identification.}
We next examine whether identification accuracy is distributed evenly across the
membership. While in the main content, we partition responding
members into a Western group and the rest, and compare accuracy within each
group under both anonymization settings. However, it is worth noticing that the 'EU', a major player within the western bloc, appears 228 times as a respondent in our corpus. A natural concern arises whether the 'EU' actually \textit{influences} the disparity in the models' behaviors. To isolate this cause, we conduct an ablation study where we purposefully discard all samples where the 'EU' is the actual respondent. This exercise results in a corpus of size 942 concerns. To this end, we investigate the disparity and present the result in \Cref{tab:western_nonwestern}.The pattern is consistent and striking: every model identifies Western respondents more accurately than non-Western ones, in both settings, with gaps ranging from roughly 5 to 14 percentage points. A chi-square test finds the disparity significant ($p<0.05$) in all but one of the twelve model--setting combinations. Crucially, the gap does not shrink when all participant identities are masked (ST-2) rather than only the respondent's (ST-1); for several models it is essentially unchanged. The ablation exercise confirms the models' geopolitical bias towards the western blocs.

\paragraph{Effect of prior negotiation history.}
A central question raised by \tradeverse{} is whether models actually use the
longitudinal structure of a concern, or merely respond to its final exchange.
To answer this directly, we conduct an ablation on the statement-generation
task. In the full setting, a model generates the respondent's final statement
given the entire preceding history---every prior meeting, together with the
raiser and supporter statements of the final round. In the ablated setting, all
prior meetings are withheld, and the model sees only the final round. Because
both settings are evaluated on the same concerns, the two conditions form
matched pairs, and any difference in generation quality can be attributed to the
presence of history rather than to differences in concern difficulty. We measure
the change in BERTScore-$F_1$ between the two settings and assess it with a
paired Wilcoxon signed-rank test, the paired effect size (Cohen's $d$), and a
95\% bootstrap confidence interval on the mean gain. As shown in
Table~\ref{tab:ablation}, the ablation results offer a useful diagnostic on model behavior. Five of six models show a significant improvement (Cohen's d between 0.32 and 0.75) when given access to full negotiation history versus the final round alone, which is suggestive that the historic context in our data carries meaningful signal that these models are able to exploit to varying degrees. GPT-OSS-120B is the one exception, showing no measurable effect (95\% CI includes zero, d = 0.05, p = 0.31).

One plausible reading is that this null result is more consistent with GPT-OSS-120B's comparatively weaker general performance on the task than with any deficiency in the underlying data's longitudinal structure, though we are cautious about drawing this conclusion too firmly from a single model's ablation profile. We note that this model trails the others across most metrics in Table 5, and its Spearman correlation between round count and BERTScore-F1 in Table 6 is the weakest of the six ($\rho$ = 0.070), reaching significance only at the 5\% level rather than the 1\% level achieved by the remaining models. This pattern is at least consistent with the possibility that a model with a weaker general grasp of the task may also be less able to productively use additional historical context, even where that context is genuinely informative. We would stop short of claiming this fully rules out other explanations — such as architectural or training differences specific to GPT-OSS-120B — and would suggest this as a hypothesis for further investigation rather than a settled conclusion.

\subsection{Experiments}
\label{app:prompts}
We report the exact prompt templates used for each task below. In each template, the placeholder \texttt{{transcript}} is replaced at inference time with the formatted meeting history of a trade concern. The history renders every round in chronological order; each statement is prefixed with the speaker's role (raiser, supporter, or respondent) and, except under the masking settings of Task~2, the member's name. For Task~2, member names are replaced by placeholder tokens (\texttt{[Country~1]}, \texttt{[Country~2]}, \dots) as described in the main paper; for Tasks~1 and~3 the transcript is unmasked.
\begin{tcolorbox}[
    colback=orange!5!white,
    colframe=orange!80!black,
    coltitle=white,
    fonttitle=\bfseries,
    title={Prompt for Task 1: HS Code Prediction},
    boxrule=0.8pt,
    arc=4pt,
    left=6pt,right=6pt,top=6pt,bottom=6pt
]
\scriptsize

You are an expert in world trade and trade-related concerns.

You are given a longitudinal series of talks.

Your task is to analyze the conversations and predict a set of appropriate Harmonized System (HS) codes.

\vspace{3pt}

\textbf{Conversation}

\textcolor{gray}{\{meeting transcript\}}

\vspace{3pt}

\textbf{Task \& Formatting Rules}

\begin{itemize}
\item You do not have to explain. 
\item Output only the HS chapter codes (e.g., \texttt{[01,02,72]}).
\item Just write the main chapter code.

\end{itemize}

\end{tcolorbox}

\begin{tcolorbox}[
    colback=green!5!white,
    colframe=green!60!black,
    coltitle=white,
    fonttitle=\bfseries,
    title={Prompt for Task 2A: Respondent Prediction (Only Respondent Masked)},
    boxrule=0.8pt,
    arc=4pt,
    left=6pt,right=6pt,top=6pt,bottom=6pt
]
\scriptsize

You are an expert in WTO trade disputes and member country regulations.

Below is a transcript of WTO meeting statements where only the responding country's name is masked as \texttt{[Country 1]}.

\vspace{3pt}

\textbf{Transcript}

\textcolor{gray}{\{meeting transcript\}}

\vspace{3pt}

\textbf{Task \& Formatting Rules}

Analyze the trade measures, policies, products, dates, and raising members to predict the true respondent WTO Member country.

\begin{itemize}
\item Output \textbf{only} the official WTO Member country.
\item Do not output reasoning, labels, punctuation, or quotation marks.
\end{itemize}

\end{tcolorbox}

\begin{tcolorbox}[
    colback=blue!5!white,
    colframe=blue!70!black,
    coltitle=white,
    fonttitle=\bfseries,
    title={Prompt for Task 2B: Respondent Prediction (All Countries Masked)},
    boxrule=0.8pt,
    arc=4pt,
    left=6pt,right=6pt,top=0pt,bottom=6pt
]
\scriptsize

You are an expert in WTO trade disputes and member country regulations.

Below is a transcript of WTO meeting statements where \textbf{all country names} are masked as placeholders like \texttt{[Country 1]}, \texttt{[Country 2]}, etc. The respondent country is represented as \texttt{[Country 1]}.

\vspace{3pt}

\textbf{Transcript}

\textcolor{gray}{\{meeting transcript\}}

\vspace{3pt}

\textbf{Task \& Formatting Rules}

Analyze the specific trade measures, tariff bound rates, presidential decrees, product sectors, dates, and legal arguments to determine the true WTO Member country represented by \texttt{[Country 1]}.

\begin{itemize}
\item Output \textbf{only} the official WTO Member country name.
\item Do not output reasoning, labels, punctuation, or quotation marks.
\end{itemize}

\end{tcolorbox}

\begin{tcolorbox}[
    colback=yellow!5!white,
    colframe=yellow!60!black,
    coltitle=white,
    fonttitle=\bfseries,
    title={Prompt for Task 3: Statement generation},
    boxrule=0.8pt,
    arc=4pt,
    left=6pt,right=6pt,top=6pt,bottom=6pt
]
\scriptsize

You are a representative of {respondent}.

You are participating in the trade negotiation.

You are given a longitudinal series of talks.

Your task is to analyze the past conversations and prepare a response for the latest conversation.

*** Past conversation ***

\textbf{Transcript} (Past coversation)

\vspace{3pt}

*** Current conversation ***

% \textbf{Transcript}

\textcolor{gray}{\{last round transcript\}}

\vspace{3pt}

\textbf{Task \& Formatting Rules}

Your response must be in the following format.

statement:YOUR RESPONSES

\end{tcolorbox}

\begin{figure*}[htbp]
\centering
\begin{tcolorbox}[colback=blue!5!white,
    colframe=blue!70!black,
    coltitle=white,
    fonttitle=\bfseries,
    title={Full dialogue for a representative concern (T\"urkiye --- EV tariffs, ID~100)},
    boxrule=0.8pt,
    arc=4pt,
    left=6pt,right=6pt,top=6pt,bottom=6pt
]
\scriptsize

\textbf{Meeting 1 --- Formal Meeting of 26--27 April 2023}\\[3pt]
\textit{China (raiser):} On 3 March, without explanation, T\"urkiye sharply increased import tariffs on China-made electric vehicles. China believes this to be inconsistent with the WTO rules. First, the relevant measure violates Article~II of the GATT~1994. According to T\"urkiye's tariff commitment, the bound rate for electric vehicles is 20\%. The import tariff of China-made electric vehicles has reached 50\%, significantly exceeding T\"urkiye's tariff commitment. Second, the relevant measure seriously violates the WTO MFN principle. T\"urkiye's action targeted only China-made electric vehicles, making the treatment of Chinese products significantly lower than that of similar products produced by other Members, which constitutes discrimination against China-made electric vehicles. China urges T\"urkiye immediately to correct its wrongdoing on China-made electric vehicles.\\[5pt]
\textit{T\"urkiye (respondent):} We would like to thank China for its interest in this issue, which is currently being discussed by our colleagues in respective Capitals. We have also taken note of the comments made here today, which will be duly conveyed. At this point, we would like to state that, as the Permanent Mission of T\"urkiye to the WTO, we have been, and will remain, open to any request from China for an exchange of opinion on this topic.

\medskip
\textbf{Meeting 2 --- Formal Meeting of 16--17 October 2023}\\[3pt]
\textit{China (raiser):} China regrets to raise this issue again. On 2 March, T\"urkiye issued a presidential decree imposing a 40\% additional tariff only on imported electric vehicles originating from China. Until now, despite various efforts made by China through both bilateral and multilateral channels, there is little movement from T\"urkiye to resolve this issue. The electric vehicles originating from China are still subject to the discriminatory tariff treatment, significantly affecting China's legitimate rights and interests under the WTO. China reiterates that the presidential decree, which imposes a 40\% additional tariff only on China-made electric vehicles, is a measure inconsistent with, among others, Articles~I and~II of the GATT~1994, as well as T\"urkiye's commitment under the WTO Agreements. In the Committee's meeting in April, the delegate of T\"urkiye stated that Capital was discussing this issue and China's comments would be conveyed. We look forward to the update from T\"urkiye today. We urge T\"urkiye to remove the discriminatory tariff without delay and bring the measures into conformity with WTO rules.\\[5pt]
\textit{T\"urkiye (respondent):} The electric vehicles sector, as an infant industry, has a strategic importance for T\"urkiye. This sector presents potentials for technological and know-how spillover to other sectors as well. Within this framework, as a result of an import surge in this sector, T\"urkiye, within its bound rates, increased its MFN tariff rate on electric vehicles for non-preferential trade partners in July~2022. Despite the introduction of the increased MFN tariff rates, an import surge continued, predominantly from China among the non-preferential trade partners. Having said that, T\"urkiye has no intention of discriminating against any Member. We took note of the concerns expressed by China today. We will provide additional information once we receive any further instructions.

\medskip
\textbf{Meeting 3 --- Formal Meeting of 25--26 March 2024}\\[3pt]
\textit{China (raiser):} In March~2023, without explanation, T\"urkiye sharply increased import tariffs on Chinese-made electric vehicles. China believes this is inconsistent with WTO rules. China is highly concerned about this measure. First, the relevant measure violates Article~II of the GATT~1994. According to T\"urkiye's tariff commitment, the bound rate for electric vehicles is 20\%. The import tariff of Chinese-made electric vehicles has reached 50\%, significantly exceeding T\"urkiye's tariff commitment. Second, the relevant measure seriously violates the WTO MFN principle. T\"urkiye's action targeted only Chinese-made electric vehicles, making the treatment of Chinese products significantly lower than that of similar products produced by other Members, which constitutes discrimination against Chinese-made electric vehicles. China urges T\"urkiye to correct its wrongdoing on Chinese-made electric vehicles immediately.\\[5pt]
\textit{T\"urkiye (respondent):} We would like to thank China for their continued interest in this issue. As indicated previously, the electric vehicles sector, as an infant industry, has a strategic importance for T\"urkiye. This sector shows potential for technological and know-how spillover to other sectors as well. T\"urkiye has no intention of discriminating against any Member. We have conveyed China's concerns to our Capital, and the Capitals are already in touch as well. We will inform our Chinese counterparts as well as the Committee on any possible development in this matter as soon as possible.

\end{tcolorbox}
\caption{The complete three-meeting dialogue for the concern shown abridged. All statements are verbatim from the WTO proceedings.}
\label{fig:full-example}
\end{figure*}
\end{document}